\documentclass[manuscript, table, nonacm]{acmart}
\usepackage{amsmath}
\usepackage{verbatim}
\usepackage{latexsym}
\usepackage{setspace}
\usepackage{mathtools}
\usepackage{bbm}
\usepackage{todonotes}
\usepackage{tikz}
\usetikzlibrary{arrows,shapes}
\usepackage{pgfplots}
\newtheorem{Definition}{Definition}
\newtheorem{Proposition}{Proposition}
\newtheorem{problem*}{Problem}
\newtheorem{Lemma}{Lemma}
\newtheorem{Theorem}{Theorem}
\newtheorem{Corollary}{Corollary}
\newtheorem{Remark}{Remark}
\newtheorem{assumption}{Assumption}

\begin{document}
\title{Event-Based Communication in Distributed Q-Learning}
\author{Daniel Jarne Ornia}
\email{d.jarneornia@tudelft.nl}
\affiliation{%
  \institution{DCSC, Delft University of Technology}
  \streetaddress{Mekelweg 2}
  \city{Delft}
  \state{The Netherlands}
  \postcode{2628CD}
}
\author{Manuel Mazo Jr.}
\email{m.mazo@tudelft.nl}
\affiliation{%
  \institution{DCSC, Delft University of Technology}
  \streetaddress{Mekelweg 2}
  \city{Delft}
  \state{The Netherlands}
  \postcode{2628CD}
}
\begin{abstract}
We present an approach to reduce the communication of information needed on a Distributed Q-Learning system inspired by Event Triggered Control (ETC) techniques. We consider a baseline scenario of a distributed $Q-$learning problem on a Markov Decision Process (MDP). Following an event-based approach, $N$ agents explore the MDP and communicate experiences to a central learner only when necessary, which performs updates of the actor $Q$ functions. We design an Event Based distributed Q learning system (\emph{EBd-Q}), and derive convergence guarantees with respect to a vanilla $Q-$learning algorithm. We present experimental results showing that event-based communication results in a substantial reduction of data transmission rates in such distributed systems. Additionally, we discuss what effects (desired and undesired) these event-based approaches have on the learning processes studied, and how they can be applied to more complex multi-agent systems.
\end{abstract}
\maketitle
\thispagestyle{empty}
\section{Introduction}
Over the past couple of decades, the interest in Reinforcement Learning (RL) techniques as a solution to all kinds of stochastic problems has exploded. In most cases, such techniques are applied to reward-maximizing problems, where an actor needs to learn a (sub) optimal policy that maximizes a time-discounted reward for any initial state. Specifically, when there is no dynamical model for the system or game, RL has proven extremely effective at finding optimal value functions that enable the construction of policies \cite{bellman2015applied,sutton2018reinforcement} maximizing the expected reward over a time horizon. This has been done with convergence guarantees for different value function forms, one of the most common ones being $Q$-Learning \cite{watkins1992q,jaakkola1994convergence}, and recently using neural networks as effective approximators of $Q$ functions \cite{mnih2015human,mnih2013playing,lillicrap2015continuous,van2016deep}. 

When the problems considered have a multi-agent nature, multi-agent theory can be combined with RL techniques as $Q$-Learning \cite{boutilier1996planning,hu1998multiagent}. In problems where a set of agents needs to optimize a (possibly shared) cost function through a model-free approach, this has been addressed in the form of Distributed $Q$-Learning \cite{weiss1995distributed,nair2015massively,horgan2018distributed,kapturowski2018recurrent}. Solutions often result in learning some form of shared policy (or value function) based on the trajectories and rewards of all agents. These techniques have been applied to many forms of competitive or collaborative problems \cite{tan1993multi,yang2004multiagent,nowe2012game,busoniu2008comprehensive,lowe2017multi}.  In the latter, agents are allowed to collaborate to reach higher reward solutions compared to a selfish approach \cite{tan1993multi,lauer2000algorithm}. This opens relevant questions regarding \emph{how} and \emph{when} to collaborate. In model free multi-agent systems, collaboration is often defined as either sharing experiences, value functions or policies, or some form of communication including current state variables of the agents. However, as it has been pointed out before \cite{kok2004sparse,panait2005cooperative}, such collaborative learning systems often include aggressive assumptions about communication between agents. Approaches to reduce this communication are, in the framework of federated learning \cite{konevcny2016federated}, in RL using efficient policy gradient methods \cite{chen2018communication}, limiting the amount of agents (or information) that communicate \cite{kok2004sparse} or allowing agents to learn how to communicate \cite{foerster2016learning,li2019graph}. These approaches focus on transmitting ``simplified" data, or modifying or learning graph topologies for the communication network.

For the problem of when to communicate over a given network, one can take inspiration from control theory approaches. When dealing with networks of sensors and actuators stabilizing a system, event-triggered control (ETC) has been established in the past decade as a technique to make networked systems more efficient while retaining stability guarantees \cite{tabuada2007event}. When applied over a distributed network, ETC allows sensor and actuator to estimate, through trigger functions, when is it necessary to communicate state samples or update controllers \cite{mazo2011decentralized,mazo2008event}. These concepts have been applied for efficient distributed stochastic algorithms \cite{george2020distributed}, or to learn parameters of linear models \cite{solowjow2020event}. In multi-agent settings, they have also been investigated to reduce the number of interactions between agents \cite{becker2004decentralized}, or to speed up distributed policy gradient methods \cite{lin2019communication,9084352}.
\subsection{Main Contribution}
Drawing a parallelism with a networked system, in this work we take inspiration from ETC techniques and turn the communication of a distributed Q-Learning problem event-based, this problem understood as in \cite{nair2015massively}, with the goal of reducing communication events, data transmission, data storage and learning steps for a fixed communication network topology. This complicates formal analysis of optimality since applying ETC-inspired trigger rules for communication effectively introduces biases in the data sampling and learning. We provide convergence guarantees of the $Q$ functions to the optimal fixed point when agents follow carefully designed fully distributed trigger functions to decide when to communicate samples. Additionally, we consider the case where we accept a certain error threshold in the optimality of the obtained $Q$ functions, and its effect on the convergence guarantees of the stochastic learning process. To the best of our knowledge, such ideas have not been applied before to distributed Q-Learning with convergence guarantees and formal bounds on optimality.

To this end, we consider a simple form of a distributed learning system that generalises many robotic learning problems: a Markov Decision Process in which agents explore to maximize the utility of a common policy in which the learning steps are centrally computed and updated periodically. In this case, explorers communicate state variables to the central learner, and the learner communicates value functions back to the explorers. We design \emph{decentralised} event triggering functions that enable such a system to \emph{maintain convergence guarantees} while each agent decides independently when to transmit experiences to the central learner over a fixed network topology. Additionally, we analyse experimentally how such event-based techniques may result in a more efficient learning process.
\section{Preliminaries}
We use calligraphic letters for sets and regular letters for functions $f:\mathbb{R}^m\to\mathbb{R}^n$. A function $f: \mathbb{R}_+\to \mathbb{R}_+$ is in class $\mathcal{K}_\infty$ if it is continuous, monotonically increasing and $f(0)=0,\,\,\lim_{a\to\infty}f(a) = \infty$. We use $\mathcal{F}$ as the measurable algebra (set) of events in a probability space, and $P$ as a probability function $P:\mathcal{F}\to [0,1]$. We use $E[\cdot]$ and $\operatorname{Var}[\cdot]$ for the expected value and the variance of a random variable. We define the set of probability vectors of size $n$ as $\mathbb{P}^n$ satisfying $p\in\mathbb{P}^n\Leftrightarrow p\geq \mathbf{0}$,  $\sum_{i=1}^n p_i =1$. Similarly, we define the set of probability matrices of size $n\times n$ as $\mathbb{P}^{n\times n}$ where $\forall i$, $\sum_{j=1}^n P_{ij}=1$. We use $\|\cdot\|_\infty$ as the sup-norm, $|\cdot |$ as the absolute value of a scalar or the cardinality of a set. We say a random process $X_n$ converges to a random variable $X$ \emph{almost surely} (a.s.) as $t\to\infty$ if it does so with probability one for any event $\omega\in\mathcal{F}$.

\subsection{MDPs and Q-Learning}
We introduce here the framework for this work. 
\begin{Definition}\label{def:MDP}[Markov Decision Process]
A Markov Decision Process (MDP) is a tuple $(\mathcal{S},\mathcal{A},P,r)$ where $\mathcal{S}$ is a set of states, $\mathcal{A}$ is a set of actions, $P: \mathcal{A}\to \mathbb{P}^{|\mathcal{S}|\times|\mathcal{S}|}$ is the probability measure of the transitions between states and $r:\mathcal{S}\times \mathcal{A}\to\mathbb{R}$ is the reward for $s\in\mathcal{S}, a\in\mathcal{A}$. 
\end{Definition}
In general, $\mathcal{S},\mathcal{A}$ are finite sets. We refer to $s,a$ as the state-action pair at time $t$, and $s,s'$ as two consecutive states. We write $P_{ss'}(a)$ as the probability of transitioning from $s$ to $s'$ when taking action $a$. 

We denote in this work a \emph{stochastic transition} MDP as the general MDP presented in Definition \ref{def:MDP}, and a \emph{deterministic transition} MDP as the particular case where the transition probabilities additionally satisfy $P_{ss'}(a)\in \{0,1\}$ (in other words, transitions are deterministic for a pair $(s,a)$).

The main goal of a Reinforcement Learning problem is to find an optimal policy $\pi^*:\mathcal{S}\to \mathcal{A}$ that maximizes the expectation of the temporal discounted reward $E[\sum_{t=0}^{\infty}\gamma^t r(s_t,a_t)\, | \pi,s_0]$ $\forall s_0 \in \mathcal{S}$ for a given discount $\gamma\in (0,1)$. To do this, we can use $Q$-Learning for the agent to learn the values of specific state-action pairs. Let the value of a state $s$ under policy $\pi$ be $ V^\pi (s) := r(s,\pi(s)) + \gamma \sum_{s'}P_{ss'}(\pi(s))V^\pi (s')$ \cite{watkins1992q}. 
The optimal value function satisfies $V^*(s): =\max_{a} r(s,a)+ \max_a \sum_{s'} P_{ss'}(a)\gamma V^* (s').$ Now define the $Q$-Values of a state-action pair under policy $\pi$ as $Q^\pi (s,a) := r(s,a) + \gamma \sum_{s}P_{ss'}(\pi(s))V^\pi (s').$ The goal of $Q$-Learning is to approximate the optimal $Q^*(s,a)$ values, which satisfy
\begin{equation}\label{eq:optQ}
Q^*(s,a): =r(s,a)+ \sum_{s'} P_{ss'}(a) \gamma V^* (s'),
\end{equation}
and yield the optimal policy $\pi^*(s):=\operatorname{argmax}_a Q^*(s,a)$ maximizing the discounted reward. For this, the $Q$-values are initialised to some value $Q_{0} (s,a) \in \mathbb{R}$ $\forall s,a$, and are updated after each transition observation $s\to s'$ with some learning rate $\alpha_t\in (0,1)$ as 
\begin{equation}\label{eq:Q}\begin{aligned}
Q_{t+1} (s,a) &= Q_{t} (s,a) +\alpha_t  \Delta_t (u),
\end{aligned}
\end{equation}
for a given sample $u=(s,a,r(s,a),s')$ and $\Delta_t (u):= r(s,a) + \gamma\max_{a'}Q_{t}(s',a') -Q_t(s,a)$ is the temporal difference (TD) error. The subscript $t$ represents the number of iterations in \eqref{eq:Q}.
\begin{Remark}
In practice, the coefficients $\alpha_t$ depend on each $(s,a)$. For ease of notation we omit this dependence, and write $\alpha_t\equiv \alpha_t (s,a)$.
\end{Remark}
The iteration on \eqref{eq:Q} is known to converge to the optimal $Q^*$ function under specific conditions of boundedness of rewards and quadratic sum convergence for the rates $\alpha_t$ (Q-Learning Convergence \cite{watkins1992q}).

\subsection{Distributed $Q$-Learning}
Let us now consider the case where $N$ agents (actors) perform exploration on parallel instances of the same MDP, with a central learner entity, generalized as an MDP with $(\mathcal{S}^N,\mathcal{A}^N,P,r)$. 

\begin{figure}
\centering
\includegraphics[width=0.4\linewidth]{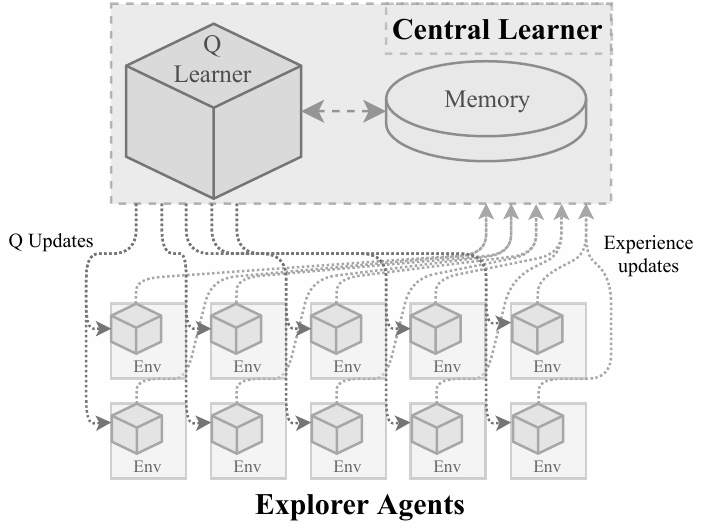}
\caption{Distributed Q-Learning System}\label{fig:2}
\end{figure}

We focus now on a \emph{distributed} $Q$-Learning system optimizing the discounted reward sum on an MDP. The goal of the distributed nature is to speed up exploration, and ultimately find the optimal policy $\pi^*$ faster. Such a system may have different architectures regarding the amount of learner entities, parameter sharing between them, \emph{etc}. We consider here a simple architecture. In this architecture, $N$ actors gather experiences of the form $u_i=(s,a,r(s,a),s')_i$ following (possibly different) policies $\pi_i$. These actors send the experiences to a single central learner, where these are sampled in batches to perform gradient descent steps on a single $\hat{Q}$ estimator, and updates each agent's policy $\pi_i$ if needed. This approach is a typical architecture on distributed $Q$-Learning problems where exploring is much less computationally expensive than learning \cite{nair2015massively}.
\begin{Definition}\label{def:decQ}
A \emph{distributed $Q$-learning} system (d-Q) for an MDP $(\mathcal{S}^N,\mathcal{A}^N,P,r)$ is a set of actor agents $\mathcal{N}=\{1,2,...,N\}$ exploring transitions and initialised at the same $s_0\in\mathcal{S}$, together with a single central learner agent storing a $\hat{Q}:\mathcal{S}\times\mathcal{A}\to\mathbb{R}$ estimator function. Let the subsets $\mathcal{N}_s=\{i\in\mathcal{N}:s_i = s\}$ have cardinality $N_s$. At each time-step, the estimator function is updated with samples $u_i$ from all $i\in\mathcal{N}$ as:
\begin{equation}\label{eq:iterdec}\begin{aligned}
\hat{Q}_{t+1}(s,a) = \hat{Q}_{t}(s,a) + \alpha_t \frac{1}{N_s}\sum_{i\in\mathcal{N}_s} \hat{\Delta}_t (u_i).
\end{aligned}
\end{equation}
\end{Definition}
We consider the following Assumption to ensure persistent exploration.
\begin{assumption}\label{as:policy}
Any policy $\pi_i$ used by an actor $i\in\mathcal{N}$ has a minimum probability $\varepsilon$ of picking an action at random.
\end{assumption}
It is straight-forward to show that the distributed form of the $Q$-Learning algorithm in \eqref{eq:iterdec} converges to the optimal $Q^*$ with probability one under the same assumptions as in Theorem 1 in \cite{watkins1992q}.
An example architecture of a distributed Q-Learning system is shown in Figure \ref{fig:2}.
\section{Problem Definition}
In practice, the distributed system in \eqref{eq:iterdec} implicitly assumes that actors provide their experiences at every step to the central learner that performs the iterations on the estimator $\hat{Q}$. When using large amounts of actors and exploring MDP's with large state-spaces, this can result in memory and data transmission rate requirements that scale badly with the number of actors, and can become un-manageable in size. Additionally, when the MDP has a unique initial state (or a small set thereof), the memory may become saturated with data samples that over-represent the regions of the state-space close to $s_0$. From this framework, we present the problem addressed in this work.
\begin{problem*}
For a distributed $Q$-learning system, design logic rules for the agents to decide when to communicate information to other agents (and when not to) that maintain convergence guarantees and reduce the system's communication requirements.
\end{problem*}
We consider the communication events to happen from explorers to learner and from learner to explorers (the communication network has a star topology). In such a system, the question of when is it useful to communicate with others naturally emerges. 

\section{Efficient Distributed Q-Learning}\label{sec:guarantees}
From the convergence proofs of Q-Learning, we know $\lim_{t\to\infty}\hat{Q}_t(s,a) - Q^*(s,a)= 0$ \emph{a.s.} Each explorer agent obtains samples $u_i=(s,a,r(s,a),s')$, and has an estimator function $\hat{Q}_t$. For every sample, the agent can compute the estimated loss with respect to the estimator $\hat{Q}_t$, which is an indication of how far the estimator is from the optimal $Q^*$. This suggests that, for $\beta\in(0,1)$, we can define a surrogate function for convergence certification to be a TD error tracking signal as:
\begin{equation}\label{eq:lyaps}
L_i(t+1) :=(1-\beta) L_i(t)+\beta|\hat{\Delta}_t(u_i)|,\, \forall i \in \mathcal{N},
\end{equation}
with  $L_i(0)=0$. We could now use $L$ to trigger communications analogously to the role of Lyapunov functions in ETC. The parameter $\beta$ serves as a temporal discount factor, that helps the agent track the TD error smoothly. Observe that:
\begin{equation*}
L_i(t)\geq 0 \,\,\forall\, t,i,\quad\text{and}\,\,\, L_i(t)\to 0 \Rightarrow \hat{Q}_t - Q^*\to 0.
\end{equation*}
The second property above is an equivalence only in the case that the MDP has deterministic transitions. The intuition about this surrogate function is as follows. Agents compute the error term $\hat{\Delta}_t(u_i)$ as they move through a trajectory, which gives an indication of how close their $\hat{Q}_t$ estimator is to the optimal $Q^*$. Then, they accumulate these losses in a temporal discounted sum $L_i(t)$, such that by storing only one scalar value they can estimate the cumulative loss in the recent past.

Recall that in \eqref{eq:optQ} the optimal $Q^*$ represents the maximum expected $Q$ values at every time step. Our convergence surrogate function in \eqref{eq:lyaps} computes the norm of the TD error at each time step, therefore it may not go to zero for stochastic transitions, but to a neighbourhood of zero. 
\begin{Proposition}\label{prop:limits}
Consider a distributed $Q$-Learning problem from Definition \ref{def:decQ}. For a deterministic MDP, $\hat{Q}_t \to Q^* \,\, a.s. \Leftrightarrow L_i(t)\to 0\,\, a.s.$ Else, if the MDP has stochastic transitions, $\hat{Q}_t \to Q^* \,\, a.s. \Rightarrow L_i(t)\to \mathcal{L}_0\,\, a.s.,$ where $\mathcal{L}_0=[0,l^*]$, and 
\begin{equation*}
l^*= \gamma \max_{s,a,s'}\left(E[\max_{a'}Q^*(s',a')|s,a]-\max_{a'}Q^*(s',a')\right).
\end{equation*}
\end{Proposition}
\begin{proof}
See Appendix \ref{apx:proofs}.
\end{proof}
\subsection{Event Based Communication}
Just as in decentralised ETC \cite{mazo2011decentralized} a surrogate for stability (Lyapunov function) can be employed to guide the design of communication triggers, we propose here using the distributed signals $L_i(t)$ based on the TD error of the $Q$ function estimators.
In our problem's context, the actors can be considered to be the sensors/actuators, and the central controller computes the iterations on $\hat{Q}_t$ based on the samples sent by the actors. This central controller updates everyone's control action (policy $\pi_i$). In the d-$Q$ system, the state variable to stabilize is the difference $E[\hat{Q}_t|\mathcal{F}_t] - Q^*$. The control action analogy is $v(k)=\hat{\Delta} (u)$, and applying $v(k)$ results in $\|E[\hat{Q}_{t+1}|\mathcal{F}_t] - Q^*\|_{\infty}<\|\hat{Q}_t- Q^*\|_{\infty}$ (see \cite{melo2001convergence}).

We set out to analyse now the implications of applying ETC techniques on the \emph{explorer}$\to$\emph{central learner} communication network. We divide the results in deterministic and stochastic transition MDPs. We refer to a communication event as an explorer agent sending a current sample $u_i$ to the central learner. To decide when to transmit, we propose to use triggering rules of the form
\begin{equation}\label{eq:etc2}
\theta_t(i) :=\left\{\begin{array}{l}1\quad \text{if} \,\,|\hat{\Delta}_t(u_i)|\geq \max\{\rho L_i(t),\epsilon\}\\
0\quad \text{else},
\end{array}\right.
\end{equation} 
with $\rho\in [0,1]$ and $\epsilon\geq 0$. That is, $\theta_t(i)=1$ means that agent $i$ sends the sample $u_i$ at time $t$ to the central learner. 
The event triggered rule in \eqref{eq:etc2} has an intuitive interpretation in the following way. Agents accumulate the value of $L_i(t)$ through their own trajectories in time. If the trajectories sample states that are already well represented in the current $\hat{Q}_t$ function, there is no need to transmit a new sample to the central learner. This can happen for a variety of reasons; some regions of the state-space may be well represented by a randomized initialisation of $\hat{Q}$, or some explorers may have already sampled the current trajectory often enough for the learner to approximate it. We refer to the resulting d-$Q$ system with event based communication as an event-based d-$Q$ (EBd-Q) system.
\subsection{Deterministic MDPs}\label{sec:ETE}
Without loss of generality, we consider the case where at every time step the samples $u_i$ are sent to the central learner, and the distributed $Q$-Learning iteration is applied over these $N$ samples (as opposed to, for example, accumulating samples over $T$ steps and learning over batches of size $T\times N$). Let us first define $H$ to be the operator:
\begin{equation}\label{eq:operator}
H(\hat{Q}_t(s,a)) :=\sum_{s'} P_{ss'}(a)\left( r(s,a) + \gamma \max_{a'}\hat{Q}_t (s',a')\right).
\end{equation}
The mapping $H$ is a contraction operator on the $\infty-$norm, with $H(Q^*)=Q^*$ being the only fixed point (see \cite{melo2001convergence} for the proof). Now observe, for a deterministic MDP, that the transition $(s,a)\to s'$ happens for a single $s'$, and:
\begin{equation*}
E[\hat{\Delta}_t (u)\, | \, \mathcal{F}_t] = H(\hat{Q}_t(s,a)) - \hat{Q}_t (s,a)=\hat{\Delta}_t (u).
\end{equation*}

\begin{Theorem}\label{lem:ETexp}
Let $(\mathcal{S}^N,\mathcal{A}^N,P,r)$ be a deterministic d-Q. Let the event triggering condition determining communication events be \eqref{eq:etc2}. Then, the resulting \emph{EBd-Q} system learning on the samples $\mathcal{U}_t^\theta:=\{u_i : \theta_t(i)=1\}$ converges \emph{a.s.} to a $\hat{Q}_{\epsilon}$ satisfying $\|\hat{Q}_{\epsilon}-\mathcal{Q}^*\|_{\infty}\leq f(\epsilon)$ with $f(\epsilon)\in\mathcal{K}_{\infty}$ under the same conditions as in \cite{watkins1992q}.
\end{Theorem}
\begin{proof}
See Appendix \ref{apx:proofs}.
\end{proof}
\begin{Remark}\label{rem:1}
Observe that for $\epsilon=0$ the triggering rule in \ref{eq:etc2} may result in regular (almost periodic) communications as $t\to\infty$. Setting $\epsilon>0$ implies the number of expected communication events goes to $0$ as $t\to\infty$, at the expense of $\hat{Q}_t$ converging to a neighbourhood of $Q^*$.\end{Remark}
One can show that, in the case of a deterministic MDP, convergence is also guaranteed for the case where $\alpha_t=\alpha$ is fixed. In practice, we can consider this to be the case when applying ET rules on a deterministic MDP.
\subsection{Stochastic MDP}
We now present similar results to Theorem \ref{lem:ETexp} for general stochastic transition MDPs. Consider a distributed MDP as in Definition \ref{def:decQ}. Let $H_P$ be the operator $H(\hat{Q}_t(s,a))$ as a function of the probability transition function $P$. Let $\mathcal{P}$ be the set of all possible transition functions for a given set of actions and states, \emph{i.e.} $\mathcal{P}\equiv \left(\mathbb{P}^{|\mathcal{S}|\times|\mathcal{S}|}\right)^{|\mathcal{A}|}$. Define $\mathcal{T}:=2^{|\mathcal{S}|\times|\mathcal{A}|\times|\mathcal{S}|}$ as the power set of transitions for the given states and actions $\mathcal{S},\mathcal{A}$. Let $G:\mathcal{T}\times\mathcal{P}\to \mathcal{P}$ be a mapping such that given a set of transitions $\tau\in\mathcal{T}$ and a transition function $P$ sets the probability of all transitions $\tau$ to zero and normalizes the resulting function $G(\tau,P)=\hat{P}^\tau\in\mathcal{P}$. 
 
Given the MDP probability measure $P$, we define the set $\mathcal{P}_P \subset \mathcal{P}$ as the set containing all transition functions $\hat{P}$ resulting from ``deleting" any combination of transitions in $P$. That is, $\mathcal{P}_P:=\{\hat{P}^\tau\}_{\tau\in\mathcal{T}}.$
Consider now the case where we apply an event triggered rule to transmit samples on a stochastic MDP. For any pair $(s,a)$, agent $i$ and time $t$, the samples are transmitted (and learned) if $\theta_t(i)=1$. This means that, in general, it may happen that for the set of resulting states $\mathcal{S}'(s,a):=\{s'\in\mathcal{S}:P_{ss'}(a)>0\}$, some transitions will not be transmitted. In practice this is equivalent to applying different transition functions $\hat{P}^\tau\in\mathcal{P}_P$ at every  step $t$. This leads to the next assumption.
\begin{assumption}\label{as:v}
There exists probability measure on $\mathcal{P}_P$, $v:\mathcal{P}_P\to [0,1]$ (or $v\in\mathbb{P}^{|\mathcal{P}_P|}$) that is only a function of the MDP $(\mathcal{S}^N,\mathcal{A}^N,P,r)$, the initial conditions $s_0, \,\hat{Q}_{0}$ and the parameters $\gamma,\varepsilon,\rho,\epsilon$, such that $v_{\hat{P}^\tau}$ is the probability of applying function $\hat{P}^\tau$ at any time step.
\end{assumption}
\begin{Remark}\label{rem:eps}
In fact it follows from the Definition of $\mathcal{P}_P$ that the dependence on $\rho,\epsilon$ must exist, given that $\rho,\epsilon= 0\Rightarrow v_{P} =1 $: in this case all samples are always transmitted. In a similar way, it also holds that $\lim_{\epsilon\to\infty}v_{\hat{P}^\tau}=0$ $\forall \hat{P}^\tau\in\mathcal{P}_P$ since in such case no samples are ever transmitted.
\end{Remark}
Let us reflect on the implications of Assumption \ref{as:v}. When applying an event triggered rule in \eqref{eq:etc2} to transmit samples, it may result on experiences not being transmitted if the trigger condition is not met. In practice, this can be modelled by considering different transition functions $\hat{P}\in\mathcal{P}_P$ (which have some values $\hat{P}^\tau_{ss'}(a)=0$ compared to the original function $P$) applied at every time-step by every agent. What Assumption \ref{as:v} implies is that, even though every agent uses different functions at every time-step, the probability of using each particular $\hat{P}^\tau\in\mathcal{P}_P$ is measurable (and stationary for fixed initial conditions).
\begin{Remark}Assumption \ref{as:v} is necessary to obtain the convergence guarantees presented in the following results. Based on the experimental results obtained, and on the fact that agents follow on-policy trajectories which are (on average) similar for the same exploration rate $\varepsilon$, the Assumption seems to hold in the cases explored. However, we leave this as a conjecture, with the possibility that the assumption could be relaxed to a time-varying distribution over $\mathcal{P}_P$.
\end{Remark}
Let us now define the operator $\tilde{H}$ as $\tilde{H}(\hat{Q}_t(s,a)):=\sum_{\hat{P}\in\mathcal{P}_P}v_{\hat{P}}H_{\hat{P}}(\hat{Q}_t(s,a))$, and let
\begin{equation}\label{eq:phi}
\Phi_t(s,a):=\frac{1}{N_s}\sum_{i\in\mathcal{N}_s}r(s,a)+\gamma\max_{a'}\hat{Q}_t(s_i',a').
\end{equation}
For a transition function $P$, set $\mathcal{P}_P$, and density $v$, define $\tilde{P}(s,a):=\sum_{\hat{P}}v_{\hat{P}}\hat{P}_{ss'}(a) $. Then, we can derive the following results.
\begin{Lemma}\label{prop:operators}
For a given agent $i$ and time $t$ transmitting samples according to the triggering condition \eqref{eq:etc2}, it holds that $E[\Phi_t(s,a)\,|\,\mathcal{F}_t] =\tilde{H}(\hat{Q}_t(s,a)),$
and the operator has a fixed point $\tilde{H}(\tilde{Q})=\tilde{Q}$ satisfying:
\begin{equation*}
\tilde{Q}(s,a):=\sum_{s'}\tilde{P}_{ss'}(a)\left( r(s,a) + \gamma \max_{a'}\tilde{Q} (s',a')\right).
\end{equation*}
\end{Lemma}
\begin{proof}
See Appendix \ref{apx:proofs}.
\end{proof}
Therefore, applying the operator $\tilde{H}(\tilde{Q})$ is equivalent to applying the contractive operator defined in \eqref{eq:operator} with a transition function $\tilde{P}$.
\begin{Theorem}\label{lem:ET2exp}
Consider a d-Q system as in Definition \ref{def:decQ}. Let the event triggering condition determining communication events be \eqref{eq:etc2}. Then, the resulting \emph{EBd-Q} system learning on the samples $\mathcal{U}_t^\theta:=\{u_i : \theta_t(i)=1\}$ converges \emph{a.s.} to a fixed point $\tilde{Q}$ under the same conditions as in \cite{watkins1992q}.
\end{Theorem}
\begin{proof}
See Appendix \ref{apx:proofs}.
\end{proof}
From Remark \ref{rem:eps} we know that ${\rho,\epsilon=0}\Rightarrow v(P) =1 \Rightarrow \tilde{Q}=Q^*$. Additionally, for $\hat{P}$ being the probability transition function applied at time $t$, it holds that $E[\hat{P}]=\tilde{P}$. But we can say something more about how the difference $P-\tilde{P}$ influences the distance between the fixed points $\|Q^*-\tilde{Q}\|_{\infty}$.
\begin{Corollary}\label{cor:1}
Let a distributed MDP with an event triggered condition as defined in \eqref{eq:etc2}. For a given transition function $P$, a set of functions $\mathcal{P}_P$ and density $v$, $\exists c\geq 0:$ 
$\|Q^*-\tilde{Q}\|_{\infty}\leq c\frac{\gamma}{1-\gamma}\|P-\tilde{P}\|_{\infty}.$
\end{Corollary}
\begin{proof}
See Appendix \ref{apx:proofs}.
\end{proof}
In fact, the distance $\|P-\tilde{P}\|_{\infty}$ is explicitly related to the probability measure $v$, since $v$ determines how far $\tilde{P}$ is from the original $P$ based on the influence of every function in the set $\mathcal{P}_P$. One can show that $\|P-\tilde{P}\|_{\infty}\leq (1-v_P)|\mathcal{P}_P|$, and $1-v_P$ is a measure of how often we use transition functions different to $P$, which depends on the aggresivity of the parameters $\rho,\epsilon$.
We continue now to study experimentally the behaviour of the Event Based d-Q systems in Theorem \ref{lem:ETexp} and \ref{lem:ET2exp} regarding the communication rates and performance of the policies obtained for a given path planning MDP problem.

\section{Experiments}\label{sec:exp}
\begin{figure}[t!]
     \centering
         \includegraphics[width=0.38\linewidth]{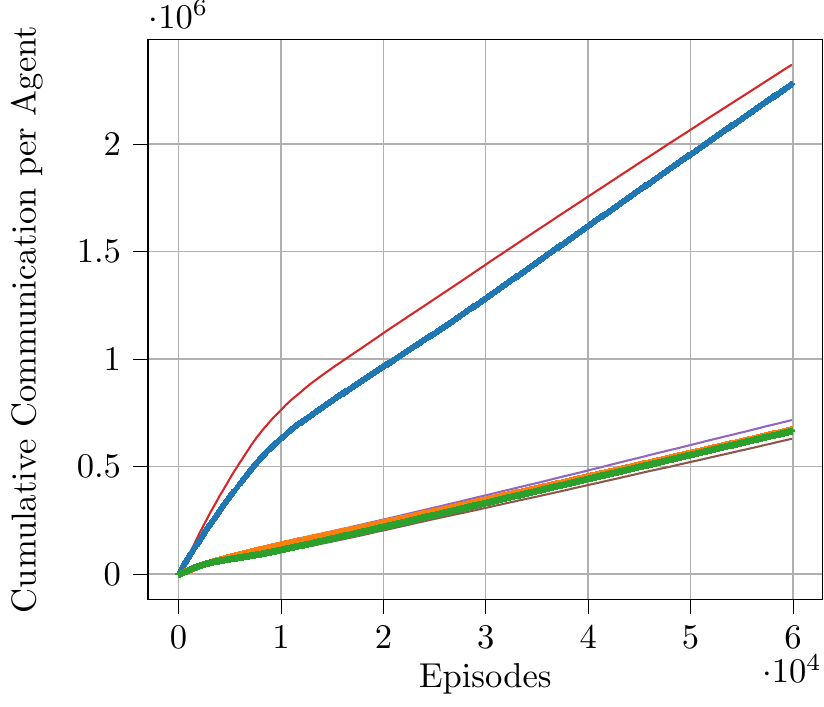}
         \includegraphics[width=0.38\linewidth]{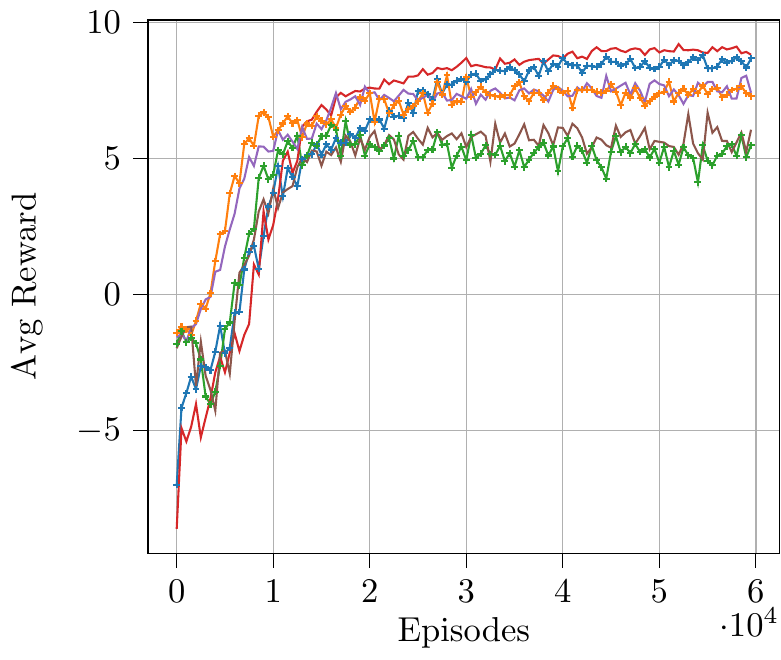}
              \includegraphics[width=0.8\linewidth]{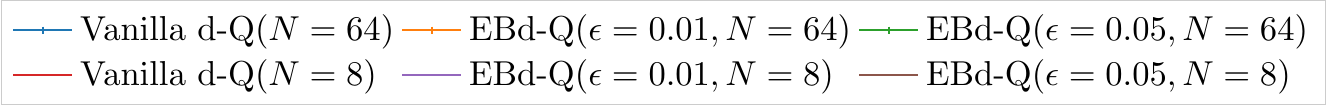}
        \caption{Stochastic Path Planning learning, Vanilla vs. EBd-Q}
         \label{fig:restoc}
\end{figure}

\begin{figure}[t!]
     \centering
         \includegraphics[width=0.38\linewidth]{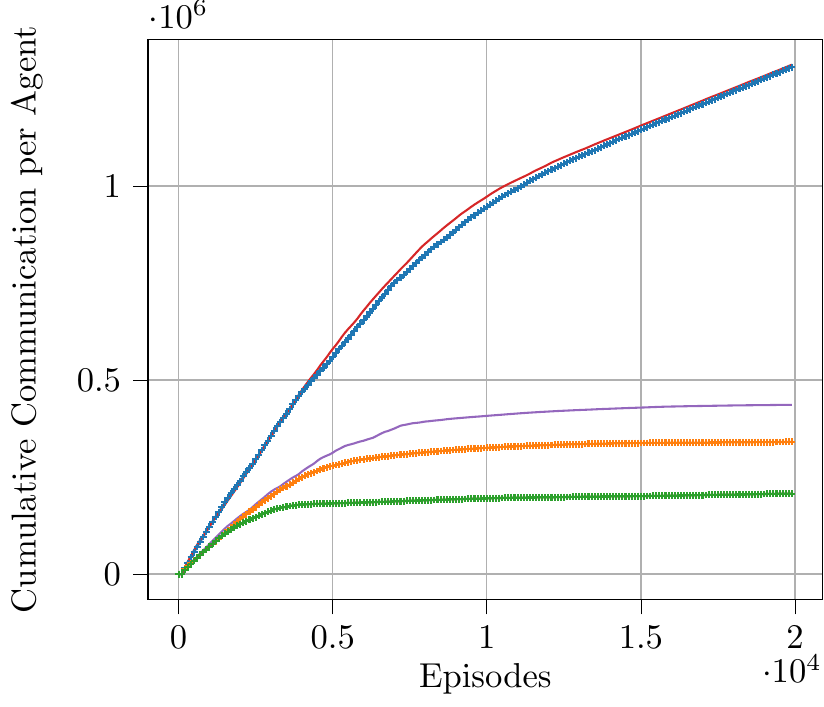}
         \includegraphics[width=0.38\linewidth]{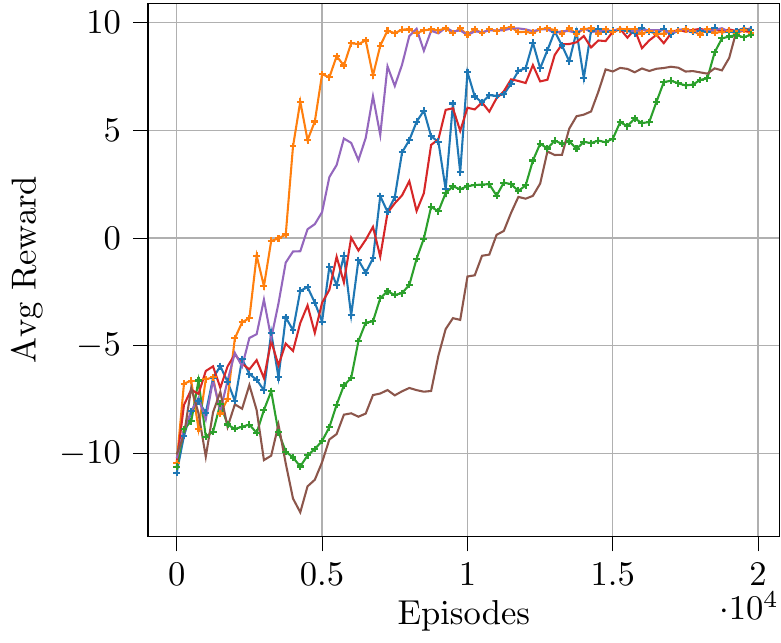}
              \includegraphics[width=0.8\linewidth]{legend.pdf}
\caption{Deterministic Path Planning learning, Vanilla vs. EBd-Q}
        \label{fig:redeter}
\end{figure}

To demonstrate the effectiveness of the different triggering functions and how they affect the learning of $Q$-values over an MDP, we use a benchmark problem consisting of a path planning problem. Details on the experimental framework are found in Appendix \ref{sec:apxex}. 
The average reward and communication results for a stochastic and deterministic MDP are presented in Figures \ref{fig:restoc} and \ref{fig:redeter}. 
We use as a benchmark a ``vanilla" distributed Q-learning algorithm where all agents are communicating samples continuously, and we compare with different combinations of parameters for the presented \emph{EBd-Q} systems. Comparing with other available research is not straight-forward, since it would require interpreting similar methods designed for other problems (in the case of distributed stochastic gradient descent works \cite{george2020distributed}, or policy gradient examples \cite{lin2019communication,9084352}), or comparing with other methods designed for learning speed (e.g. \cite{kapturowski2018recurrent}), where the goal is not to save communication bandwidth or storage capacity.

Analysing the experimental results, in both the stochastic and deterministic MDP scenarios, the systems reach an optimal policy quicker by following an event triggered sample communication strategy, but only for $\epsilon=0.01$. This can be explained by the same principle as in \emph{prioritized sampling} \cite{schaul2015prioritized,horgan2018distributed}: samples of un-explored regions of the environment are transmitted (and learned) earlier and more often. However, in our case this emerges as a consequence of the trigger functions $\theta_i(t)$, and it is the result of a fully distributed decision process where agents decide independently of each-other when to share information, and does not require to accumulate and sort the experiences in the first place. When increasing the triggering threshold to $\epsilon=0.05$, the learning gets compromised and the reward decreases for both $N=64$ and $N=8$.

Additionally, we observe in both scenarios how the total number of communications increase much slower in the event based case compared to the vanilla d-$Q$ example, and even stabilize in the case of the deterministic MDP, indicating the number of events is approaching zero. This is due to the \emph{EBd-Q} systems sending a much lower amount of samples through the network per time step.

At last, as anticipated by the theoretical results in Theorems \ref{lem:ETexp} and \ref{lem:ET2exp}, higher $\epsilon$ results in a larger reduction of communication rates, at the expense of obtaining less optimal $Q$ functions.
\section{Discussion}
We have presented a design of ETC inspired trigger functions for RL agents to determine when to share experiences with a central learner on a distributed MDP. Additionally, we derived convergence guarantees for both a general stochastic transition MDP and for the particular case where the transitions in the MDP are deterministic. The goal is to allow agents in a d-Q system to make distributed decisions on which particular experiences may be valuable and which ones not, reducing the amount of communication events (and data transmission and storage). 

Regarding the convergence guarantees, we have shown how applying such triggering functions on the communication events results in the centralised learner converging to a $Q$-Function that may slightly deviate from the optimal $Q^*$. However, we were able to provide an indication on how far the resulting $Q$ functions can be from $Q^*$ based on the triggering parameters $\epsilon,\rho$, explicitly for a deterministic MDP and implicitly (via the distribution $v$) for a stochastic MDP. Interesting observations arise from both the theoretical and the experimental results. Event based rules appear to reduce significantly the amount of communication required in the explored path planning problem, while keeping a reasonable learning speed. In fact, it was observed in the experiments how the proposed \emph{EBd-Q} systems resulted collaterally in a faster learning rate than for the constant communication case (an effect similar to that in \emph{prioritized learning}).

Finally, some questions for future work emerge from these results. First, the study of event-based $\hat{Q}$ function updating from \emph{central learner}$\to$\emph{explorers}. Likewise, it would be interesting to explore the effect of such event based communication on general multi-agent RL systems where all agents are learners, the communication graph has a complete topology, and agents could be sharing more than experiences ($Q-$values, policies...). Such study would shine light on how to design efficient collaborative multi agent systems. At last, we leave as a conjecture whether Assumption \ref{as:v} always holds, left for future work, with the possibility of analysing \emph{EBd-Q} systems as some form of alternating or interval MDP where the transition functions belong in some set. 
\section*{Acknowledgements}
The authors want to thank G. Delimpaltadakis, G. Gleizer and M. Suau for the useful discussions. This work is partly supported by the ERC Starting Grant SENTIENT 755953.
\bibliographystyle{ACM-Reference-Format}
\bibliography{biblo}
\clearpage
\appendix
\section{Technical Proofs}\label{apx:proofs}
\begin{proof}[Proposition \ref{prop:limits}]
Consider first a deterministic MDP. In this case, $P_{ss'}(a)\in\{0,1\}$, and $Q^*(s,a)=r(s,a) + \gamma \max_{a'}Q^*(s',a')$. Then, it must hold
\begin{equation*}
\hat{Q}_t \to Q^*\,\,a.s. \Leftrightarrow \hat{Q}_t(s,a)-Q^*(s,a) \to 0\,\,a.s. \,\, \,\,\forall s,a. 
\end{equation*}
Recalling the $Q-$learning iteration, $\forall s,a$ it holds \emph{almost surely}:
\begin{equation}\label{eq:fixedV}\begin{aligned}
&\lim_{t\to\infty}\hat{Q}_t(s,a)-Q^*(s,a) = 0\Leftrightarrow \\
\Leftrightarrow&\lim_{t\to\infty}\hat{Q}_t(s,a)-r(s,a) - \gamma \max_{a'}\hat{Q}_t(s',a')= 0 \Leftrightarrow\\
\Leftrightarrow & \lim_{t\to\infty}\hat{\Delta}_t(u)= 0.
\end{aligned}
\end{equation}
Therefore, $\hat{\Delta}_t(u)\to 0 \Leftrightarrow |\hat{\Delta}_t(u)|\to 0$, and $\lim_{t\to\infty}L(t)=0 \Leftrightarrow \lim_{t\to\infty}|\hat{\Delta}_t(u)|=0$, which happens almost surely.

For the stochastic transition MDP, the optimal $Q^*$ function satisfies
\begin{equation*}
Q^*(s,a) = r(s,a)+\sum_{s'} P_{ss'}(a) \gamma \max_{a'}Q^*(s',a').
\end{equation*}
Therefore, similarly to \eqref{eq:fixedV}, $\forall s,a$:
\begin{equation*}\begin{aligned}
\lim_{t\to\infty}&\hat{Q}_t(s,a)-Q^*(s,a) = 0 \,\,a.s. \Leftrightarrow \\
\Leftrightarrow &\lim_{t\to\infty} \hat{\Delta}_t(u)= \gamma\max_{a'}Q^*(s',a')- \\
&-\sum_{s'} P_{ss'}(a)\gamma \max_{a'}Q^*(s',a')=:\hat{\Delta}^*(u)\,\, a.s.
\end{aligned}
\end{equation*}
At last, from Assumption \ref{as:policy}, every pair $(s,a)$ is visited infinitely often. Therefore, 
\begin{equation*}\begin{aligned}
\|\hat{\Delta}^*(u)\|_\infty=&\gamma \|E[\max_{a'}Q^*(s',a')|s,a]-\max_{a'}Q^*(s',a')\|_\infty:=l^*,
\end{aligned}
\end{equation*}
and $L(t+1)\leq (1-\beta)L(t)+\beta l^*\Rightarrow \lim_{t\to\infty}L(t) \leq l^* $ \emph{a.s.}
\end{proof}
\begin{proof}[Theorem \ref{lem:ETexp}]
We show this by contradiction. Assume first that $\exists t_0$ such that a communication event is never triggered after $t_0$. Then, $|\hat{\Delta}_t(u_i)|<\rho L_i(t)$ and $L_i(t+1)\leq (1-\beta(1-\rho))L_i(t)$ $\forall t>t_0$. Therefore, $\exists t_\epsilon\geq t_0:$ $L_i(t_\epsilon) < \epsilon$, and from \eqref{eq:optQ} $\forall (s,a)$ and $\forall t>t_\epsilon$:
\begin{equation}\label{eq:lyap1}\begin{aligned}
&|r(s,a) + \gamma\max_{a'}\hat{Q}_{t}(s',a') -\hat{Q}_t(s,a)|\leq \epsilon\Rightarrow \\
\Rightarrow& |Q^*(s,a) -\hat{Q}_t(s,a) + \gamma\left(\max_{a'}\hat{Q}_{t}(s',a')- \max_{a'}Q^*(s',a')\right)|\leq \epsilon\Rightarrow\\
\Rightarrow& |Q^*(s,a) -\hat{Q}_t(s,a)|\leq  \epsilon + \gamma\max_{a'}\left |\hat{Q}_{t}(s',a')- Q^*(s',a')\right|\leq \epsilon + \gamma\|\hat{Q}_t-Q^* \|_\infty\Rightarrow\\
\Rightarrow & \|Q^* -\hat{Q}_t\|_\infty\leq \epsilon + \gamma\|\hat{Q}_t-Q^* \|_\infty
\leq \frac{\epsilon}{1-\gamma},
\end{aligned}
\end{equation}
where $f(\epsilon):=\frac{\epsilon}{1-\gamma}$ is a $\mathcal{K}_\infty$ function. Furthermore, it follows from the ET condition that no samples are transmitted for $t>t_\epsilon$, therefore $\hat{Q}_t$ has converged for $t>t_{\epsilon}$ to some $\hat{Q}_{\epsilon}$. Therefore, $\lim_{t\to\infty}\|Q^* -\hat{Q}_t\|_\infty = \|Q^* -\hat{Q}_{\epsilon}\|_\infty\leq \frac{\epsilon}{1-\gamma}$.
Now assume that communication events happen infinitely often after some $t_0$. 
Since all pairs $(s,a)$ are visited infinitely often, $\hat{\Delta}_{t+1}(u) = H(\hat{Q}_{t+1})(s,a)-\hat{Q}_{t+1}(s,a)$ and
\begin{equation*}\begin{aligned}
&\|\hat{\Delta}_{t+1}(u)\|_{\infty} \leq \|H(\hat{Q}_{t})(s,a)-\hat{Q}_{t}(s,a) + \alpha_t \left(\gamma \max_{a'}\hat{\Delta}_t(u')-\hat{\Delta}_t(u) \right)\|_{\infty}\leq\\
=& \|(1-\alpha_t)\hat{\Delta}_t(u) + \alpha_t\gamma \max_{a'}\hat{\Delta}_t(u')\|_{\infty}\leq (1-\alpha_t(1-\gamma))\|\hat{\Delta}_{t}(u)\|_{\infty}.
\end{aligned}
\end{equation*}
Therefore, $\lim_{t\to\infty}\|\hat{\Delta}_{t}(u)\|_{\infty}=0$, which implies no samples are transmitted as $t\to\infty$, which contradicts the \emph{infinitely often} assumption. Therefore from \eqref{eq:lyap1}, $\lim_{t\to\infty}\|Q^* -Q_t\|_\infty\leq \frac{\epsilon}{1-\gamma}$.
\end{proof}
\begin{proof}[Lemma \ref{prop:operators}]
First, from Assumption 2, for a given $\hat{P}$, $
E[\Phi_t(s,a)\,|\,\mathcal{F}_t,\hat{P}] = H_{\hat{P}}(\hat{Q}_t)(s,a).$
Now, by the law of total expectation and making use of $\Pr[\hat{P}]=v_{\hat{P}}$, it follows that $ E[\Phi_t(s,a)\,|\,\mathcal{F}_t] = \sum_{\hat{P}\in\mathcal{P}_P}v_{\hat{P}}H_{\hat{P}}(\hat{Q}_t)(s,a).$
At last, to show that $\tilde{Q}$ is a fixed point, observe we can write
\begin{equation*}\begin{aligned}
&\sum_{\hat{P}\in\mathcal{P}_P}v_{\hat{P}}\sum_{s'} \hat{P}_{ss'}(a)\left( r(s,a) + \gamma \max_{a'}\tilde{Q} (s',a')\right)=\\
=&\sum_{s'}\left(\sum_{\hat{P}}v_{\hat{P}}\hat{P}_{ss'}(a) \right)\left( r(s,a) + \gamma \max_{a'}\tilde{Q} (s',a')\right)=\\
=&\sum_{s'}\tilde{P}_{ss'}(a)\left( r(s,a) + \gamma \max_{a'}\tilde{Q} (s',a')\right).
\end{aligned}
\end{equation*}
\end{proof}
\begin{proof}[Theorem \ref{lem:ET2exp}]
Define $\xi_t(s,a):=\hat{Q}_t(s,a)-\tilde{Q}(s,a)$. Then, the iteration \eqref{eq:Q} applied at every time step is 
\begin{equation*}
\xi_{t+1}(s,a) = (1-\alpha_t)\xi_t(s,a)+\alpha_t (\Phi_t(s,a)-\tilde{Q}(s,a)).
\end{equation*}
Now, from Lemma \ref{prop:operators}, 
\begin{equation*}\begin{aligned}
&\|E[\Phi_{t+1}(s,a)-\tilde{Q}(s,a)|\mathcal{F}_t]\|_{\infty}=\|\tilde{H}(\hat{Q}_{t+1})(s,a)-\tilde{H}(\tilde{Q})(s,a) \|_{\infty}=\\
&=\gamma\|\tilde{P}_{ss'}(a)\left( \max_{a'}\hat{Q}_t (s',a')- \max_{a'}\tilde{Q} (s',a')\right)\|_{\infty}\leq \gamma\|\tilde{P}_{ss'}(a) \|_{\infty}\|\hat{Q}_t-\tilde{Q} |\|_{\infty}\leq\gamma\|\xi_t(s,a) |\|_{\infty}.
\end{aligned}
\end{equation*}
Therefore, the expected value of the operator $\tilde{H}$ is a $\gamma$-contraction in the sup-norm, with fixed point $\tilde{Q}$, and it follows that $\|\xi_t(s,a)\|_{\infty}\to 0$ \emph{a.s}.
\end{proof}
\begin{proof}[Corollary \ref{cor:1}]
Recall $\tilde{H}(\tilde{Q})=\tilde{Q}$ and $H({Q^*})={Q^*}$. Then,
\begin{equation}\label{eq:Qdistance}\begin{aligned}
&\|Q^*-\tilde{Q}\|_{\infty}=\|H({Q^*})-\tilde{H}(\tilde{Q})\|_{\infty}=\\
=&\|\sum_{s'} P_{ss'}(a)\left( r(s,a) + \gamma \max_{a'}Q^* (s',a')\right)-\\
&-\tilde{P}_{ss'}(a)\left( r(s,a) + \gamma \max_{a'}\tilde{Q} (s',a')\right)\|_{\infty}=\\
=&\gamma\|\sum_{s'} P_{ss'}(a)\max_{a'}Q^* (s',a')-\\
&-\tilde{P}_{ss'}(a)\max_{a'}\tilde{Q} (s',a')\|_{\infty}.
\end{aligned}
\end{equation}
Define $\hat{\Delta} P_{ss'}(a) := \tilde{P}_{ss'}(a)-P_{ss'}(a)$ and substitute in \eqref{eq:Qdistance}:
\begin{equation}\label{eq:Qdistance2}\begin{aligned}
&\|Q^*-\tilde{Q}\|_{\infty}=\gamma\|\sum_{s'} P_{ss'}(a)\left(\max_{a'}Q^* (s',a')-\right.\\
&-\left.\max_{a'}\tilde{Q} (s',a')\right)-\hat{\Delta} P_{ss'}(a)\max_{a'}\tilde{Q} (s',a')\|_{\infty}\leq\\
\leq &\gamma \|\sum_{s'}P_{ss'}(a)\max_{a'}|{Q^*} (s',a')-\tilde{Q} (s',a')|\|_{\infty}+\\
&+\gamma\|\hat{\Delta} P_{ss'}(a)\max_{a'}\tilde{Q} (s',a')\|_{\infty}.
\end{aligned}
\end{equation}
At last, observe $\|\sum_{s'}P_{ss'}(a)\max_{a'}|{Q^*} (s',a')-\tilde{Q} (s',a')|\|_{\infty}\leq \gamma \|Q^*-\tilde{Q}\|_{\infty}$. Additionally, since the reward functions are bounded, for a discount rate $\gamma\in(0,1)$ the values of $\tilde{Q}(s,a)\leq c$ are also bounded for some constant $c\in\mathbb{R}_+$. Therefore,
\begin{equation}\label{eq:Qdistance3}\begin{aligned}
\|Q^*-\tilde{Q}\|_{\infty}\leq & \gamma \|Q^*-\tilde{Q}\|_{\infty} +\gamma\|\hat{\Delta} P_{ss'}(a)\|_{\infty} \|\tilde{Q}\|_{\infty}\leq\\
\leq & \gamma \|Q^*-\tilde{Q}\|_{\infty} +c\gamma\|\hat{\Delta} P_{ss'}(a)\|_{\infty} \Rightarrow\\
\Rightarrow \|Q^*-\tilde{Q}\|_{\infty}\leq& c\frac{\gamma}{1-\gamma}\|P-\tilde{P}\|_{\infty}.
\end{aligned}
\end{equation}
\end{proof}
\section{Experimental Framework}\label{sec:apxex}
All experiments were run on a MacBook Pro with 2,3 GHz Quad-Core Intel Core i5 and 8GB RAM. The path planning environment considered is very light-weight and the examples do not make use of any computationally heavy method (neural network training, \emph{etc}), therefore we were able to run all experiments on a single CPU. The agent's random exploration is implemented using Numpy's \emph{random.uniform} to decide on the $\varepsilon$-greedy policy, and \emph{random.randint} to pick a random action.

We modified the Frozen Lake environment in OpenAI GYM \cite{brockman2016openai}. We edited the environment to have a bigger state-space ($1296$ $(s,a)$ pairs), the agents get a reward $r=-1$ when choosing an action that makes them fall in a hole, and $r=10$ when they find the goal state. Additionally, the agents get a constant reward of $-0.01$ every time they take an action, to reflect the fact that shorter paths are preferred. The action set is $\mathcal{A}=\{\text{up},\,\text{down},\,\text{left},\,\text{right}\}$. For the stochastic transition case, the agents get a reward based on the pair $(s,a)$ regardless of the end state $s'$. The resulting Frozen Lake environment can be seen in Figure \ref{fig:1} in the Appendix.
We consider a population of $N\in\{8,64\}$ agents, all using $\varepsilon$ greedy policies with different exploration rates (as proposed in \cite{mnih2016asynchronous}). The number of agents is chosen to be multiple of 8 (to facilitate running on parallel cores of the computer), to represent both a ``large" and a ``small" agent number scenario. The agents are initialised with a value $\varepsilon_i\in\{0.01,0.2,0.4,0.6,0.8.0.99\}$ chosen at random. For all the simulations we use $\alpha=0.01$, $\gamma=0.97$, $\beta=0.05$ and $\rho=0.9$. We plot results for $\epsilon\in\{0.01,0.05\}$. The $Q-$function is initialised randomly $\hat{Q}_0(s,a)\in [-1,1]\, \forall\,(s,a)$. The results are computed for 25 independent runs and averaged for each scenario.
We present results for a stochastic and a deterministic MDP. In the stochastic case, for a given pair $(s,a)$ there is a probability $p=0.7$ of ending up at the corresponding state $s'$ (e.g. moving down if the action chosen is \emph{down}) and $\bar{p}=0.3$ of ending at any other adjacent state.

To compare between the different scenarios, we use an experience replay buffer of size $N\times 1000$ for the central learner's memory, where at every episode we sample mini-batches of $32$ samples. The policies are evaluated by a critic agent with a fixed $\varepsilon_0=0.01$, computing the rewards for 10 independent runs for every estimation $\hat{Q}_t$.
\begin{figure}
\centering
\includegraphics[width=0.3\linewidth]{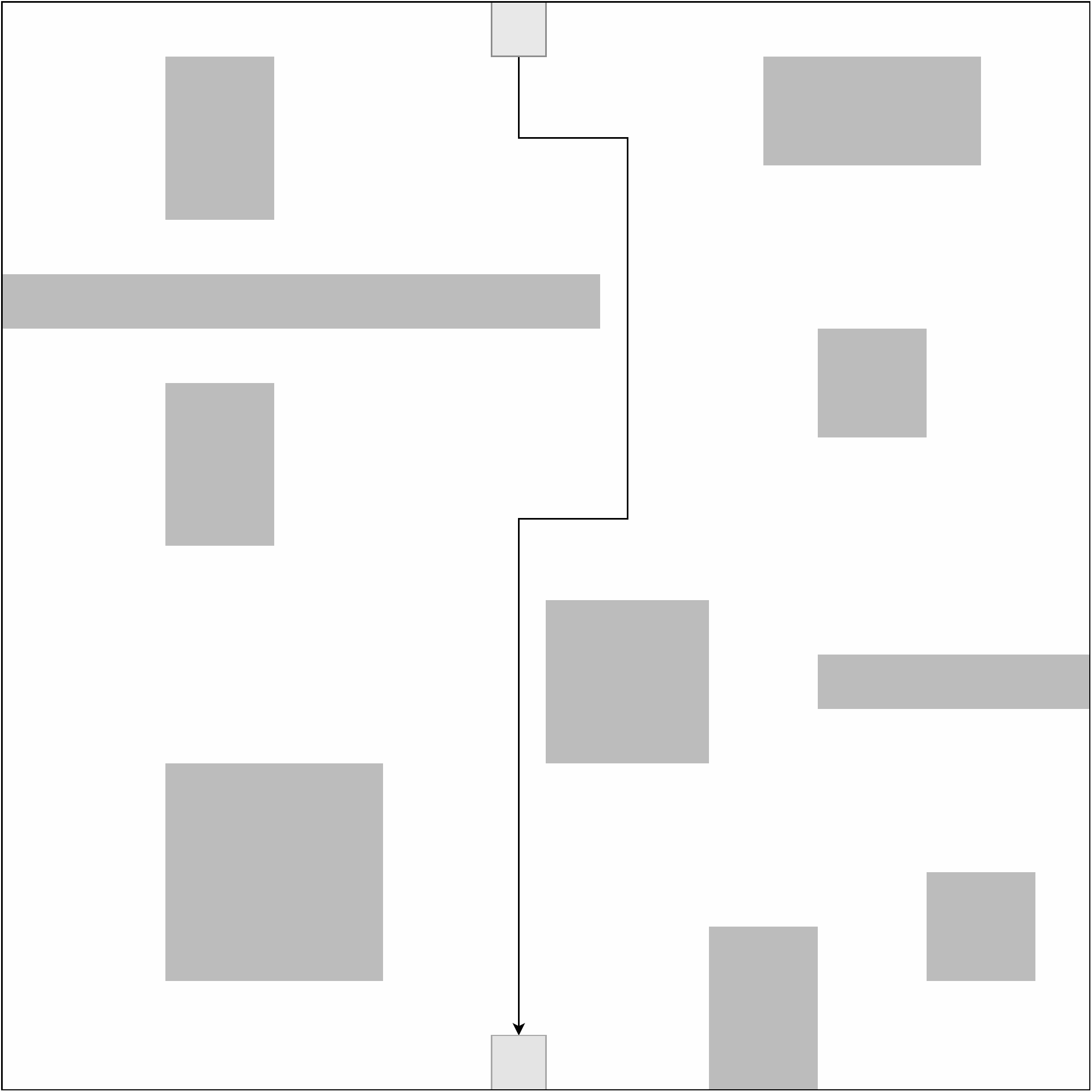}
\caption{Frozen Lake of size 18$\times$18 used for path planning experiments.}\label{fig:1}
\end{figure}
The learning rate $\alpha$ and ``diffusion" $\gamma$ were picked based on similar size $Q$-learning examples in the literature. In the case of the ET related parameters $\beta, \rho, \epsilon$, these were picked after a very light parameter scan to illustrate significant properties related to the theoretical results. First, $\beta=0.05$ yields a half-life time of $\approx 15$ time steps, which is on the same order as the diameter of the path planning arena. The value of $\rho$ was just picked arbitrarily close to 1 to allow a slow decrease in the communication rate. At last, $\epsilon$ was chosen first to be close to $0$, but keeping in mind the order of magnitude of the rewards $\|r(s,a)\|_\infty=10$. The variable $\epsilon$ acts as an error threshold, under which the errors in the $Q$ values are considered low enough and no samples are transmitted. The value function magnitude is related to the maximum reward in the MDP. As an example, a pair $(s,a)$ being 1 step away from the path planning goal has an associated reward on the order of $\gamma\|r(s,a)\|_\infty\approx 9.7$. However, a pair $(s,a)$ being 2 steps away has $\gamma^2\|r(s,a)\|_\infty\approx 9.4$. Therefore, when being really close to the goal, the error associated with taking one extra step is on the order of $\approx 0.03$. By choosing $\epsilon=0.01$, we ensure the threshold is low enough to capture one-step errors. Then, $\epsilon=0.05$ is larger than this gap, so it ensures a significant enough difference for comparison.
\begin{figure}[t!]
     \centering
         \includegraphics[width=0.49\linewidth]{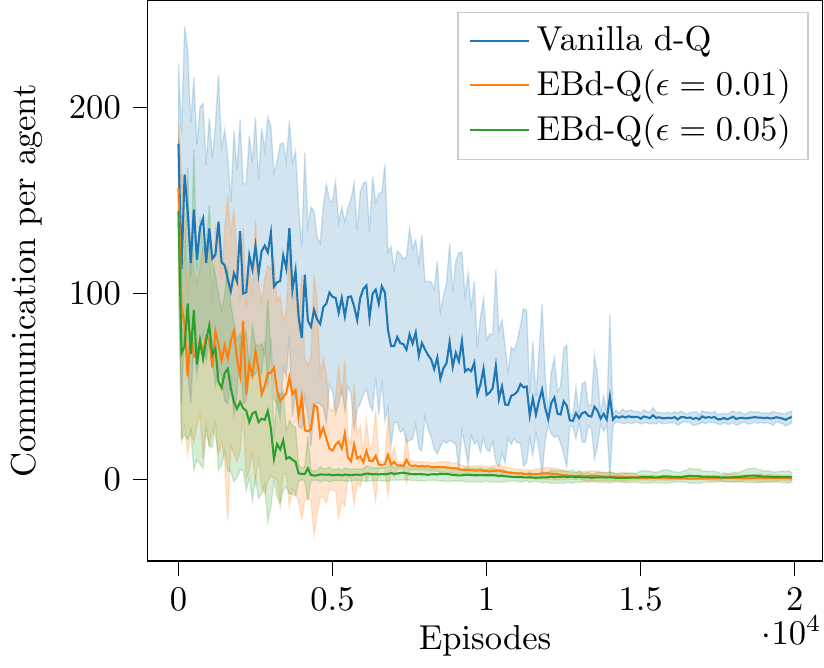}
         \includegraphics[width=0.49\linewidth]{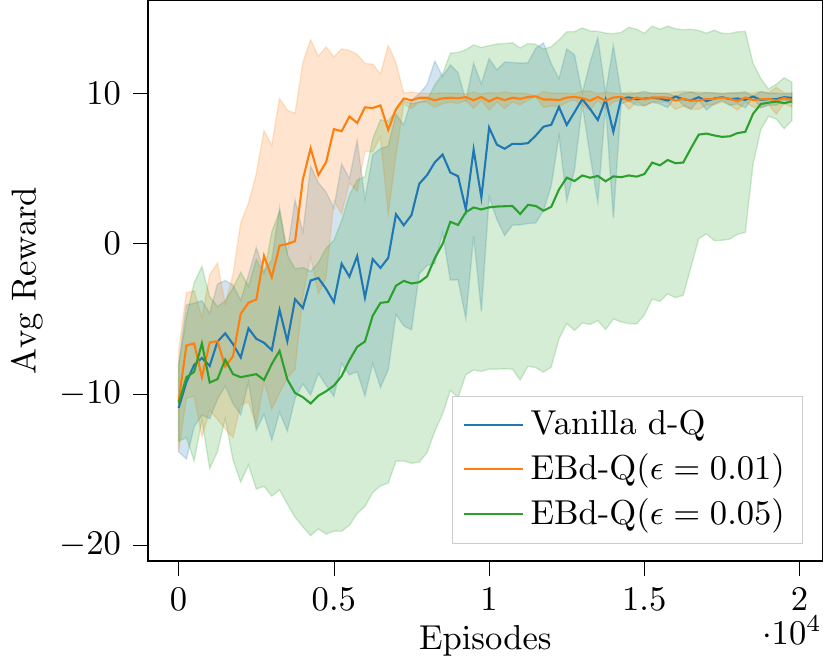}
\caption{Deterministic Path Planning Results with Variance for $N=64$}
        \label{fig:redeter_full}
\end{figure}
\begin{figure}[t!]
     \centering
         \includegraphics[width=0.49\linewidth]{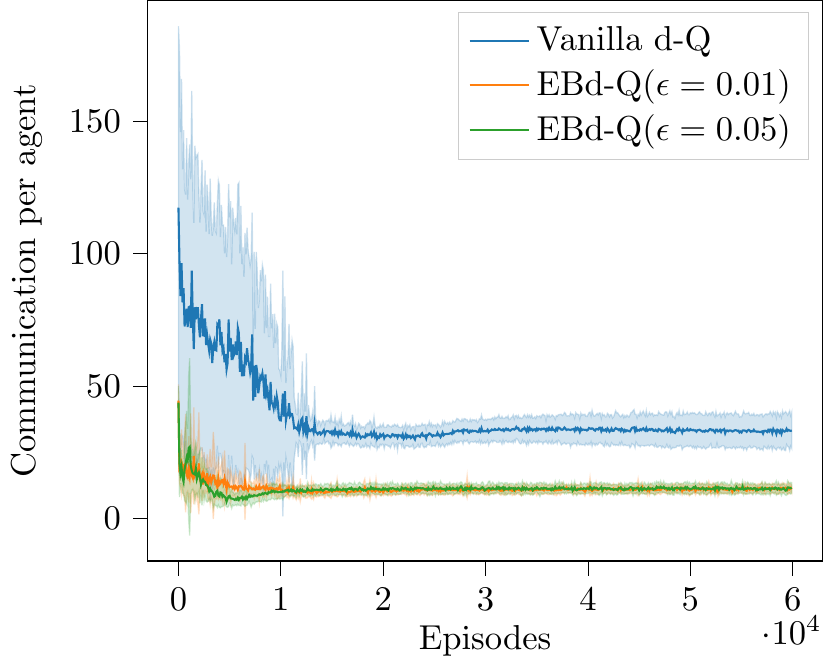}
         \includegraphics[width=0.49\linewidth]{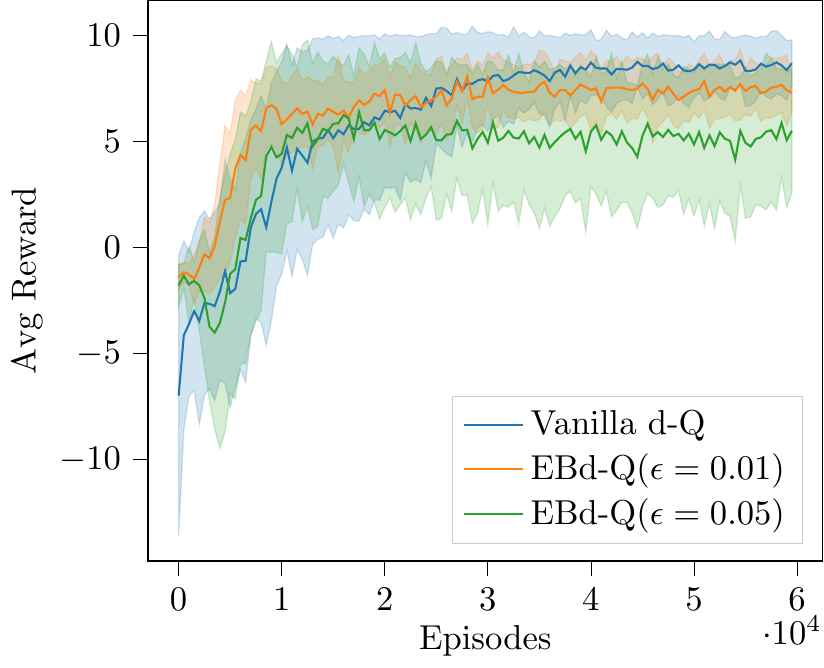}
\caption{Stochastic Path Planning Results with Variance for $N=64$}
        \label{fig:restoc_full}
\end{figure}
Figures \ref{fig:redeter_full},\ref{fig:restoc_full} show the results with the standard deviation for the three scenarios with $N=64$. The critic agent was limited to $1500$ steps when evaluating the policies, to speed up the policy evaluation, since it was the case for the deterministic MDP that the agent would learn to not fall in a hole without finding the goal, taking extremely large numbers of steps to evaluate a single policy. In general, it is the case that periodic communication patterns result in smaller variances in both updates and reward values. Additionally, one can see how on the deterministic MDP case, the variances go practically to zero as soon as optimal policies have been found.
\end{document}